\pdfoutput=1

\documentclass[11pt]{article}

\usepackage[final]{acl}
\usepackage{tablefootnote}
\usepackage{times}
\usepackage{latexsym}
\usepackage{threeparttable}
\usepackage{amsmath}
\usepackage{booktabs}
\usepackage{multirow}
\usepackage{arydshln}
\usepackage[T1]{fontenc}
\usepackage{soul} 
\usepackage{pifont}
\newcommand{\cmark}{\ding{51}} 
\newcommand{\xmark}{\ding{55}} 

\usepackage{color, xcolor}

\usepackage[utf8]{inputenc}
\usepackage{enumitem}
\setenumerate[1]{itemsep=1.5pt,partopsep=0pt,parsep=\parskip,topsep=5pt}
\setitemize[1]{itemsep=1.5pt,partopsep=0pt,parsep=\parskip,topsep=5pt}
\usepackage{microtype}

\usepackage{inconsolata}

\usepackage{graphicx}
\title{Look \& Mark: Leveraging Radiologist Eye Fixations and Bounding boxes in Multimodal Large Language Models for Chest X-ray Report Generation}

\author{
    \textbf{Yunsoo Kim}$^{1}$ \quad
    \textbf{Jinge Wu}$^{1}$ \quad
    \textbf{Su-Hwan Kim}$^{2}$ \\
    \textbf{Pardeep Vasudev}$^{1}$ \quad
    \textbf{Jiashu Shen}$^{3}$ \quad
    \textbf{Honghan Wu}$^{1,4}$ \\
    $^1$UCL \quad
    $^2$Technical University of Munich \quad
    $^3$University of Oxford \quad
    $^4$University of Glasgow \\
    \texttt{\{yunsoo.kim.23, honghan.wu\}@ucl.ac.uk}
}

\begin{document}
\maketitle
\begin{abstract}
Recent advancements in multimodal Large Language Models (LLMs) have significantly enhanced the automation of medical image analysis, particularly in generating radiology reports from chest X-rays (CXR). However, these models still suffer from hallucinations and clinically significant errors, limiting their reliability in real-world applications. In this study, we propose \textbf{Look \& Mark (L\&M)}, a novel grounding fixation strategy that integrates radiologist eye fixations (Look) and bounding box annotations (Mark) into the LLM prompting framework. Unlike conventional fine-tuning, L\&M leverages in-context learning to achieve substantial performance gains without retraining. 
When evaluated across multiple domain-specific and general-purpose models, L\&M demonstrates significant gains, including a 1.2\% improvement in overall metrics (A.AVG) for CXR-LLaVA compared to baseline prompting and a remarkable 9.2\% boost for LLaVA-Med. General-purpose models also benefit from L\&M combined with in-context learning, with LLaVA-OV achieving an 87.3\% clinical average performance (C.AVG)—the highest among all models, even surpassing those explicitly trained for CXR report generation. Expert evaluations further confirm that L\&M reduces clinically significant errors (by 0.43 average errors per report), such as false predictions and omissions, enhancing both accuracy and reliability. These findings highlight L\&M's potential as a scalable and efficient solution for AI-assisted radiology, paving the way for improved diagnostic workflows in low-resource clinical settings.
\end{abstract}

\section{Introduction}
\label{sec:intro}

Recently, the advent of multimodal Large Language Models (LLMs), in which vision encoders are integrated with powerful language generation models, has significantly advanced the automation of medical image analysis \cite{li2023comprehensive,li2024llavamed,wu2023radfm,microsoftmultimodal2024,saab2024capabilities}. Chest X-ray (CXR) interpretation, in particular, has benefited greatly from these developments: by ingesting both image and text data, modern LLMs can generate radiology reports, perform visual question answering, and even conduct error-checking in clinical documents \cite{hyland2023maira,chen2024chexagent,lee2023cxr,wu2023exploring, wu2024slava}. 

Despite these advances, hallucinations, outputs diverging from actual image content, remain a major obstacle for real-world applications of these models, reducing the trust and clinical userability. \cite{xiao2024comprehensive,alsaad2024multimodal,chen2024detecting,wu2024hallucination}. One promising way to overcome these hallucinatory behavior is by incorporating expert insights directly into the model pipelines. Several studies have shown that integrating human input with AI models can substantially boost both accuracy and reliability, sometimes surpassing the performance of either clinicians or AI models alone \citep{calisto2022hitlbreast,hitlperformance}. Two relevant sources of expert knowledge in radiology are (1) bounding boxes, which are rectangular markers that radiologists draw to highlight suspicious regions in medical images, and (2) radiologist eye fixations, which reveal the natural diagnostic process by tracking where doctors look and how long they spend examining different areas of a chest X-ray.

\begin{table*}[htbp]
\centering
\footnotesize
\begin{tabular}{lccccc}
\toprule
\textbf{Method} & \textbf{Uses Gaze} & \textbf{Uses BBox} & \textbf{LLM-Based} & \textbf{Training-Free} & \textbf{Report Generation} \\
\midrule
\textbf{MAIRA2}~\cite{bannur2024maira2} & \xmark & \cmark & \cmark & \xmark & \cmark \\
\textbf{CXR Fixation}~\cite{kim2024human} & \cmark & \xmark & \cmark & \xmark & \xmark \\
\textbf{FG-CXR}~\cite{pham2024fgcxr} & \cmark & \cmark & \xmark & \xmark & \cmark \\
\textbf{Ours (Look \& Mark)} & \cmark & \cmark & \cmark & \cmark & \cmark \\
\bottomrule
\end{tabular}
\caption{Comparison of related methods. Our method is the only one that is LLM-based, training-free (prompt-only), and integrates both radiologist gaze and bounding box information for report generation.}
\label{tab:related_work_comparison}
\end{table*}

Bounding box annotations help localize the language output in well-defined spatial coordinates, reducing the risk of free-form hallucinations and improving multimodal large language model's “grounded” report generation ability \cite{bannur2024maira2}. Meanwhile, eye-tracking data offers insights into the contextual logic clinicians apply, indicating not only the spatial information but also the order of saliency for the spatial information. This additional information enhanced the capabilities of multimodal LLMs in CXR interpretations including report generation \cite{kim2024enhancing,kim2024human}. Each approach brings a complementary perspective: bounding boxes offer explicit “marks” of suspicious regions, whereas fixation data conveys their relative significance with the duration of "looking".

In this work, we propose \textbf{“Look \& Mark”}: a grounding fixation strategy for CXR report generation that merges radiologist eye fixations with bounding box annotations in multimodal LLMs. Crucially, we avoid large-scale model retraining by employing \emph{in-context learning} or prompt engineering. Through bounding box coordinates, we provide precise “marks” that ground suspicious regions, while eye fixations encode how expert radiologists spatially and temporally navigate those regions. This grounding fixation can significantly reduce hallucinations and clinically significant errors by generating more coherent, clinically relevant CXR reports.

The contributions of our work are as follows:
\begin{itemize}[leftmargin=1em,noitemsep,topsep=0pt]
    \item \textbf{Novel Integration Framework - Performance improvement Without Re-Training:} We propose a systematic approach for combining spatial (bounding boxes) and temporal (eye fixations) expert knowledge in a single unified framework, enabling more comprehensive image understanding that mirrors expert diagnostic processes without re-training for domain adaptation or task specific fine-tuning.
    
    \item \textbf{Fewer Hallucinations and Errors:} We show, through radiologist expert evaluations, that grounding fixation strengthens the alignment of generated text with the ground truth reports, mitigating one of the most pressing drawbacks of large-scale LLM-based solutions for radiology.
    
    \item \textbf{Comprehensive Evaluation Across Multiple LLMs:} We validate \textbf{Look \& Mark} on several general-purpose and medical multimodal LLMs, demonstrating consistent gains in accuracy.
\end{itemize}

\section{Related Works}
\label{sec:related_work}

\subsection{MAIRA2: Grounded Radiology Report Generation} 

MAIRA2 is a large multimodal radiology-specific model designed for grounded report generation \cite{bannur2024maira2}. The model incorporates bounding box annotations as spatial constraints, ensuring that each finding in the generated report is explicitly localized on the CXR image. By grounding language outputs in bounding boxes, MAIRA2 mitigates hallucinations and improves alignment between generated text and image content. Additionally, the model integrates contextual inputs, such as prior imaging studies and clinical indications, to further enhance report accuracy and completeness. Despite its strong performance, MAIRA2 focuses solely on bounding box grounding and does not incorporate the dynamic reasoning patterns captured through radiologists’ eye fixations. 

\subsection{Chest X-ray Diagnosis with Eye Fixation} 

Kim et al.\cite{kim2024human} explored the role of radiologists’ fixation data in guiding multimodal LLMs for CXR analysis. By incorporating fixation-based textual prompts and aligning fixations with anatomical bounding boxes, the study demonstrated improvements in classification tasks such as diagnosis and report error-checking (presence or absence). However, the study  primarily focuses on diagnostic tasks and does not address the report generation task directly. It also does not leverage bounding boxes of abnormalities. Our work addresses these gaps by combining gaze information with abnormalities’ bounding boxes to guide multimodal LLMs more effectively in generating radiology reports.

\begin{figure*}[htbp]
\centering
\includegraphics[width=\linewidth]{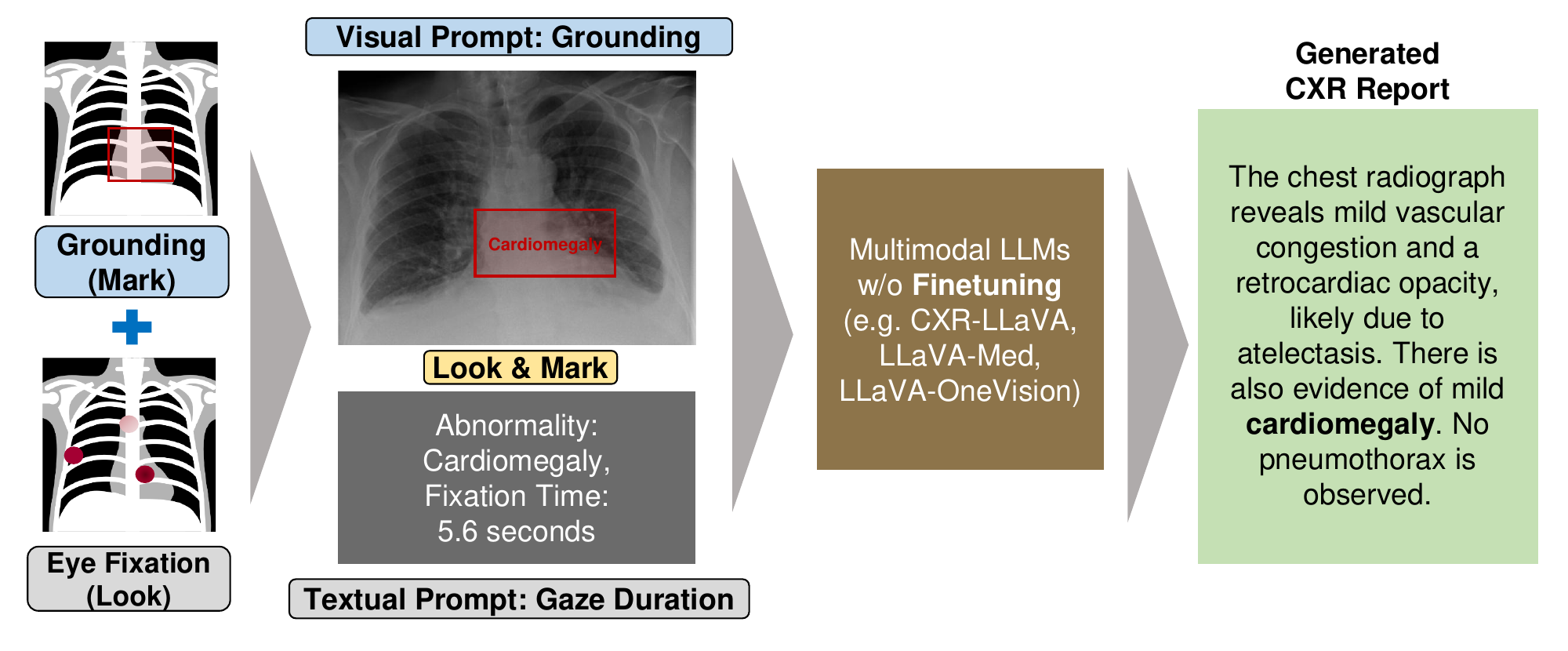}
\caption{
Overview of the \textbf{Look \& Mark} framework. The input to the model consists of a chest X-ray image augmented with two forms of expert-derived visual grounding: (1) \textbf{Bounding boxes} highlighting abnormal findings (Mark), and (2) \textbf{Radiologist eye fixations}, converted into fixation heatmaps and summarized in the prompt as text (Look). The bounding boxes are overlaid on the image as part of the visual input, while the fixation durations mapped to abnormalities are embedded in the textual prompt. These dual cues are used to construct an in-context prompt for a multimodal LLM, enabling it to generate clinically relevant, grounded radiology reports without model retraining. To comply with the MIMIC-CXR data usage license, we use a substitute image from Wikimedia that reflects a comparable diagnosis, and paraphrase the associated report text.
}
\label{fig:overview}
\end{figure*}

\subsection{FG-CXR: Fine-Grained Alignment of Gaze and Text} 

FG-CXR is a dataset, which aligns radiologist gaze Lmaps with anatomical segmentation masks and corresponding diagnostic report text \cite{pham2024fgcxr}. This dataset was used to develop the Gen-XAI framework, which generates CXR reports by leveraging gaze attention Lmaps to ground textual outputs in anatomical regions.

While FG-CXR advances interpretability in report generation, it does not employ multimodal large language models, instead relying on gaze-linked text to train a specific vision and language model. Furthermore, it lacks the use of bounding box annotations to explicitly ground abnormalities spatially. Our grounding fixation approach extends this work by unifying gaze data within abnormalities’ bounding boxes, improving the performance of report generation tasks in multimodal LLMs without requiring additional training.

\section{Look \& Mark}
\label{sec:methods}

Figure~\ref{fig:overview} provides an overview of the workflow, which includes preprocessing the input, constructing multimodal prompts, and evaluating the output.

\subsection{Abnormalities Bounding box Integration}

Let the input image be \( I \) and the bounding boxes of abnormalities be \( B = \{b_1, b_2, \dots, b_n\} \), where each bounding box \( b_i \) is defined as:
\begin{equation}
b_i = (x_{i1}, y_{i1}, x_{i2}, y_{i2}, l_i),
\end{equation}
where \((x_{i1}, y_{i1})\) and \((x_{i2}, y_{i2})\) are the top-left and bottom-right coordinates, respectively, and \( l_i \) is the associated abnormality label (e.g., ``Cardiomegaly''). For each bounding box, an abnormality caption is assigned, as shown in Figure~\ref{fig:overview}. 

\subsection{Eye Fixation Integration}

Fixation data \( G = \{g_1, g_2, \dots, g_m\} \) is represented as:
\begin{equation}
g_j = (x_j, y_j, t_j),
\end{equation}
where \((x_j, y_j)\) denotes the coordinates of the fixation point, and \( t_j \) is the duration of fixation at that location. Each fixation point is mapped to a corresponding bounding box \( b_i \) in the set of bounding boxes \( B = \{b_1, b_2, \dots, b_n\} \), which represent regions of abnormalities in the chest X-ray.

The mapping is defined as:
\begin{equation}
\mathcal{M}(g_j, B) = b_i \; \text{if} \; (x_j, y_j) \in b_i,
\end{equation}

where \( b_i = (x_{i1}, y_{i1}, x_{i2}, y_{i2}) \) defines the coordinates of the bounding box. This mapping links each fixation point to the associated abnormality label \( l_i \) of the bounding box \( b_i \). 

In cases where multiple bounding boxes contain the same fixation point, we resolve the ambiguity by selecting the smallest bounding box (i.e., the one with the minimum area). This prioritization reflects the assumption that tighter boxes offer more precise localization of abnormalities and ensures that each fixation is attributed to the most relevant region.

For each bounding box \( b_i \), we compute the total fixation time \( T_i \), which represents the cumulative duration of all fixations mapped to that bounding box:
\begin{equation}
T_i = \sum_{g_j \in G, \mathcal{M}(g_j, B) = b_i} t_j.
\end{equation}

The fixation data is formatted into textual prompts following this template:
 \texttt{Fixation Data: [Abnormality bounding box: \{label\}, Fixation Time: \{time\} seconds]}. This format encodes the temporal patterns of radiologist attention for each identified abnormality.

As Figure \ref{fig:overview} demonstrates, the \textbf{bounding box annotations} are used as a visual prompt, providing spatial guidance to the model by highlighting abnormalities directly in the image. The \textbf{fixation data} linked to abnormalities is provided as a textual prompt, encoding temporal significance and prioritization. 

\section{Experiments}
\label{sec:experiments}

\subsection{Dataset}
For this study, we used two primary datasets: the \textbf{REFLACX} dataset as the source of eye fixation data and dictated reports, and the \textbf{MS\_CXR} dataset as the source of abnormalities bounding boxes \cite{bigolin2022reflacx,boecking2022mscxr}. Both datasets are derived from the widely used \textbf{MIMIC\_CXR} dataset, which provides chest X-ray images alongside corresponding findings and impression sections of radiology reports \cite{johnson2019mimic}.

The MIMIC\_CXR dataset has served as a foundation for many radiology report generation models, which often rely heavily on the findings and impression sections for training and evaluation. However, several key challenges exist:
\begin{itemize}[leftmargin=1em,noitemsep,topsep=0pt]
    \item \textbf{Incomplete Annotations:} Some chest X-ray images in MIMIC\_CXR lack findings or impression sections, reducing the reliability of these sections as a sole reference for report generation.
    \item \textbf{Low Inter-Rater Agreement:} Interviews conducted during the design of this study revealed that even expert radiologists often disagree on chest X-ray interpretations. This variability in interpretation further questions the validity of using a single ground-truth report per image. 
    \item \textbf{Free-Form Terminology:} Radiology reports are inherently free-form in style and terminology, making it challenging to evaluate models against a single predefined ground-truth report.
\end{itemize}

To address these limitations, we chose to use the dictated reports from the REFLACX dataset rather than the standard MIMIC\_CXR findings and impressions. REFLACX provides multiple dictated reports per image, which better capture the variability in radiologist interpretation and report terminology. This approach allows for a more robust evaluation of our model's ability to generalize across diverse reporting styles.

\begin{table}[htbp]
\centering
\begin{tabular}{lcc}
\toprule
\textbf{Category}                        & \textbf{Statistic}  \\ \midrule
\textbf{General Statistics} & \\
Number of Images                        & 560      \\
Number of Dictated Reports              & 1,372    \\
Average Reports per Image               & 2.5     \\ \midrule
\textbf{Missing MIMIC\_CXR Reports} & \\
Images without Findings      & 210      \\
Images without Impression    & 129      \\ \midrule
\textbf{Average Text Lengths (Characters)} & \\
Findings Section                        & 410.3    \\
Impression Section                      & 227.5    \\
Dictated Reports                        & 243.7    \\
\bottomrule
\end{tabular}
\caption{Look \& Mark Dataset Statistics.}
\label{tab:data_statistics}
\end{table}

Furthermore, to enhance the dataset, we integrated bounding box annotations from the MS\_CXR dataset, which offers precise localization of abnormalities. These bounding boxes are linked with textual prompts and fixation durations, providing additional multimodal data that can be leveraged to analyze the relationship between image regions of interest and the corresponding dictated text.

Table \ref{tab:data_statistics} highlights the key statistics of the \textbf{Look \& Mark dataset}, a combined dataset created for this study. It includes 560 chest X-ray images paired with 1,372 dictated reports, averaging 2.45 reports per image. Notably, 210 images lack findings sections and 129 images lack impression sections in the original MIMIC\_CXR dataset, emphasizing the gaps that the Look \& Mark dataset helps to fill.

The dictated reports from the dataset are shorter (average length of 243.7 characters) than typical findings (410.3 characters) but offer free-form descriptions closely resembling real-world clinical reporting styles. Each report is also accompanied by bounding box annotations for abnormalities, along with fixation duration data, providing a unique multimodal dataset that combines textual, visual, and spatial information.

\subsection{Models}

\begin{table}[htbp]
\footnotesize
\centering
\begin{tabular}{@{}lcccc@{}}
\toprule
  \textbf{Model Name}  & \textbf{Size} & \textbf{Trained for} \\
\midrule
\textbf{CXR-LLaVA}\citep{lee2023cxr} & 8B & RG \\
\textbf{MAIRA2}\citep{bannur2024maira2} & 7B & Ground RG \\
LLaVA-Med\citep{li2024llavamed} & 8B & Medical VQA \\
 \midrule 
Llama3.2V\citep{dubey2024llama} & 11B & IU  \\
LLaVA-OV\citep{li2024llavaov} & 8B & VU \\
Qwen2.5VL\citep{Qwen2.5-VL} & 8B & Grounding, VU \\
\bottomrule
\end{tabular}
\caption{Model descriptions. Models in bold are trained with the MIMIC-CXR dataset. Acronym for the tasks are as follows: Report Generation - RG, Visual Question Answering - VQA, Image Understanding - IU, Vision Understanding (Image, Video) - VU}
\label{tab:overview_models}
\end{table}

Table \ref{tab:overview_models} describes the models evaluated in this study. The selection includes both models specifically fine-tuned for the medical domain and general-purpose multimodal models, enabling a comprehensive comparison of their performance under the Look \& Mark (L\&M) approach.

Two of the evaluated models, \textbf{CXR-LLaVA} and \textbf{MAIRA2}, were fine-tuned on the MIMIC-CXR dataset. These models are specifically designed for radiology tasks, making them well-suited for chest X-ray interpretation. \textbf{CXR-LLaVA}, with 8 billion parameters, was trained for report generation (RG) and focuses on translating visual abnormalities into detailed radiology reports. On the other hand, \textbf{MAIRA2}, built on the Mistral 7B backbone, is uniquely trained for grounded report generation (Ground RG), linking textual reports to specific regions of interest in the chest X-ray. 

In contrast, \textbf{LLaVA-Med} was trained using an instruction tuning dataset derived from figures and legends in PubMed papers. This model, with 8 billion parameters, was designed for medical visual question answering (VQA) tasks, demonstrating strong reasoning capabilities but lacking specific training on MIMIC-CXR data.

To provide broader context, the study also includes general-purpose multimodal models such as \textbf{Llama3.2V}, \textbf{LLaVA-OV}, and \textbf{Qwen2.5VL}. \textbf{Llama3.2V}, the largest model with 11 billion parameters, is trained for image understanding (IU), providing insights into how parameter scaling impacts performance. \textbf{LLaVA-OV} and \textbf{Qwen2.5VL}, both with 8 billion parameters, focus on vision understanding (VU), encompassing tasks involving image and video comprehension. \textbf{Qwen2.5VL} is the only model that is trained to do object grounding. While these models were not fine-tuned for the medical domain, they serve as baselines for evaluating generalization capabilities on CXR report generation.

\subsection{Evaluation}

To evaluate the performance of the models on the Look \& Mark dataset, we tested different input modalities and grounding strategies, including default prompts (-), eye fixation data represented as heat maps (L), bounding box grounding (M), and our proposed grounding fixation approach combining both heat maps and bounding boxes (L\&M). For general-purpose large language models (LLMs) not fine-tuned on the dataset, in-context learning (I) was applied. This involved providing three exemplar reports, each with different style of writing and not included in the dataset but sourced from REFLACX dictated reports, as context to teach the model chest X-ray writing style.

For hyperparameters, we used a batch size of 1 and a temperature of 0 or 0.1 (where 0 is not accepted). The temperature was chosen to minimize the randomness in the generated report. The maximum length of new tokens was 512 tokens.

We used both lexical and clinical relevance evaluation metrics:
\begin{itemize}[leftmargin=1em,noitemsep,topsep=0pt]
    \item \textbf{ROUGE-L \cite{lin2004rouge}:} Measures the lexical overlap between the generated and reference reports, emphasizing long matching sequences.
    \item \textbf{BERTScore \cite{zhang2019bertscore}:} Computes semantic similarity between generated and reference reports by comparing token embeddings, offering a more nuanced view of report coherence.
    \item \textbf{RadGraph-XL \cite{delbrouck2024radgraphxl}:} Evaluates the ability of models to extract clinically relevant entities and relations, assessing how well the generated reports align with annotated medical knowledge graphs.
    \item \textbf{RaTEScore \cite{zhao2024ratescore}:} A metric tailored for radiology report evaluation, emphasizing clinical entities, negations, and synonym robustness to assess the quality of generated text.
    \item \textbf{Clinical Metrics (C.AVG):} This score is calculated by normalizing each clinical relevance metric (RadGraph-XL and RaTEScore) by the highest scores, converting each to a percentage, and then averaging them. It provides a unified percentage-based metric to assess clinical utility.
    \item \textbf{All Metrics (A.AVG):} Similarly, this score is calculated by normalizing all metrics (ROUGE-L, BERTScore, RadGraph-XL, RaTEScore, and others) by their respective highest scores, converting them to percentages, and then taking the average. It provides a comprehensive, normalized view of model performance across all evaluation dimensions.
\end{itemize}

\paragraph{Clinical Metric Selection Rationale.}
While CheXpert F1~\cite{smit2020chexbert} has been widely used in past work, several recent studies have shown that it correlates poorly with expert radiologist evaluations, particularly for complex, multi-sentence report generation tasks. Instead, we adopt more recent and clinically grounded evaluation metrics such as \textbf{RadGraph-XL}~\cite{delbrouck2024radgraphxl} and \textbf{RaTEScore}~\cite{zhao2024ratescore}, which are specifically designed to assess clinical accuracy and factual consistency in radiology reports. These metrics have demonstrated stronger alignment with human expert judgment compared to earlier classification-based metrics like CheXpert F1.

\section{Results and Discussion}
\label{sec:results}

\begin{table*}[htbp]
\footnotesize
\centering
\begin{tabular}{llcccccc}
\toprule
\textbf{Model} & \textbf{Method} & \textbf{RG-L} & \textbf{BERT} & \textbf{RadG} & \textbf{RaTE} & \textbf{C.AVG (\%)} & \textbf{A.AVG (\%)} \\ 
\midrule
CXR-LLaVA & - & 0.1653 & 0.8586 & 0.1107 & 0.4730 & 84.42 & 73.21 \\
\multirow{2}{*}{\textbf{CXR-LLaVA}} & \multirow{2}{*}{\textbf{L\&M}} & \textbf{0.1697} & \textbf{0.8602} & 0.1148 & 0.4752 & 86.01 & \textbf{74.40} \\
& &  (+0.0044) & (+0.0016) & (+0.0041) & (+0.0022) & (+1.59) & (+1.19) \\ \hdashline
MAIRA2 & - & 0.1460 & 0.8492 & 0.0868 & 0.4507 & 74.35 & 66.70 \\
\multirow{2}{*}{MAIRA2} & \multirow{2}{*}{L\&M} & 0.1469 & 0.8489 & 0.0810 & 0.4574 & 73.16 & 66.31 \\
& &  (+0.0009) & (-0.0003) & (-0.0058) & (+0.0067) & (-1.19) & (-0.39) \\ \hdashline
LLaVA-Med & - & 0.0942 & 0.8392 & 0.0000 & 0.2445 & 24.99 & 40.62 \\
\multirow{2}{*}{LLaVA-Med} & \multirow{2}{*}{L\&M} & 0.0817 & 0.8253 & 0.0295 & 0.4191 & 52.46 & 49.81 \\
 & &  (-0.0125) & (-0.0139) & (+0.0295) & (+0.1746) & (+27.47) & (+9.19) \\
\hdashline
Llama3.2V & - & 0.0413 & 0.7652 & 0.1412 & 0.0027 & 46.34 & 41.32 \\
\multirow{2}{*}{Llama3.2V} & \multirow{2}{*}{L\&M} & 0.0393 & 0.7694 & 0.1494 & 0.0071 & 49.44 & 42.30 \\
& & (-0.0020) & (+0.0042) & (+0.0082) & (+0.0044) & (+3.10) & (+0.98) \\
\multirow{2}{*}{\textbf{Llama3.2V}} & \multirow{2}{*}{\textbf{I\&L\&M}} & 0.0402 & 0.7696 & \textbf{0.1533} & 0.0089 & 50.91 & 42.99 \\
& & (-0.0011) & (+0.0044) & (+0.0121) & (+0.0062) & (+4.57) & (+1.67) \\ \hdashline
LLaVA-OV & - & 0.0518 & 0.8085 & 0.0471 & 0.3936 & 55.58 & 47.14 \\
\multirow{2}{*}{LLaVA-OV} & \multirow{2}{*}{L\&M} & 0.0527 & 0.8107 & 0.0497 & 0.4531 & 62.51 & 50.07 \\
& & (+0.0009) & (+0.0022) & (+0.0026) & (+0.0595) & (+6.93) & (+2.93) \\
\multirow{2}{*}{\textbf{LLaVA-OV}} & \multirow{2}{*}{\textbf{I\&L\&M}} & 0.0959 & 0.8365 & 0.1145 & \textbf{0.4893} & \textbf{87.34} & 65.69 \\
& & (+0.0441) & (+0.0280) & (+0.0674) & (+0.0957) & (+31.76) & (+18.55) \\ \hdashline
Qwen2.5VL & - & 0.0576 & 0.8080 & 0.0534 & 0.4291 & 61.27 & 50.08 \\
\multirow{2}{*}{Qwen2.5VL} & \multirow{2}{*}{L\&M} & 0.0427 & 0.7933 & 0.0528 & 0.4488 & 63.08 & 48.71 \\
& & (-0.0149) & (-0.0147) & (-0.0006) & (+0.0197) & (+1.81) & (-1.37) \\
\multirow{2}{*}{Qwen2.5VL} & \multirow{2}{*}{I\&L\&M} & 0.0614 & 0.8045 & 0.0812 & 0.4730 & 74.83 & 55.88 \\
& & (+0.0038) & (-0.0035) & (+0.0278) & (+0.0439) & (+13.56) & (+5.80) \\ 
\bottomrule
\end{tabular}
\caption{Performance for all the models using L\&M and I\&L\&M compared to baseline (-). Numbers in parentheses on the second row for each method indicate the absolute difference from the baseline method for the same model. RG-L: ROUGE-L, BERT: BERTScore, RadG: RadGraph-XL, RaTE: RaTEScore. The best scores for each metric are bolded.}
\label{tab:all_model_key_metrics}
\end{table*}

\subsection{Performance Comparison Across Models and Methods}

Table~\ref{tab:all_model_key_metrics} presents a comprehensive evaluation of model performance across key metrics: ROUGE-L (RG-L), BERTScore, RadGraph-XL (RadG), and RaTEScore (RaTE). These results demonstrate the effectiveness of \textbf{Grounding Fixation Prompting (L\&M)} in improving report generation performance across both domain-specific and general-purpose models. Furthermore, the extension of L\&M with in-context learning, denoted as \textbf{I\&L\&M}, significantly enhances general-purpose models’ adaptability.

\subsubsection{Domain-Specific Models.}
Among domain-specific models, \textbf{CXR-LLaVA} demonstrates the highest improvement when incorporating the L\&M strategy. For instance, RG-L increases from 0.1653 (default prompting) to 0.1697, and BERTScore improves from 0.8586 to 0.8602. Clinical average improves from 84.42\% to 86.01\%, indicating that L\&M aligns better with clinical expectations as well. These improvements highlight the ability of grounding fixation to enhance both the linguistic and clinical quality of generated reports.

However, in the case of \textbf{MAIRA2}, the results reveal a more nuanced outcome. While L\&M slightly improves RG-L (0.1469 vs. 0.1460), there is a small decline in C.AVG (73.16\% vs. 74.35\%). This suggests that MAIRA2's architecture may already effectively integrate bounding box information, leaving limited room for further enhancement with gaze data. Additionally, the architectural complexity or pretraining objectives of MAIRA2 might not optimally benefit from the added eye fixation cues. 

For \textbf{LLaVA-Med}, we see a huge performance boost with L\&M in clinical relevance evaluation metrics, while decreased performance in the lexical evaluation metrics. As the decresae in lexical evaluation metrics was marginal when compared to the performance boost in the clinical evaluation metrics, the overal average score (A.AVG) resulted in 9.19\% increase.

\subsubsection{General-Purpose Models}
General-purpose models, \textbf{Llama3.2V}, \textbf{LLaVA-OV}, and \textbf{Qwen2.5VL}, also experience in performance improvement with L\&M. LLaVA-OV performance increased in all evaluation metrics. Llama3.2V performance increased in all evaluation metrics except ROUGE-L. However, Qwen2.5VL model only increased in RaTEScore. The performance increase in these general domain models, which have not been specifically trained with chest X-ray datasets, adds the generalizability of the L\&M prompting.

They also benefit significantly from the addition of in-context learning (I) combined with L\&M. For \textbf{LLaVA-OV}, I\&L\&M achieves notable improvements across all metrics, with BERTScore increasing to 0.8365 and RaTEScore to 0.4893. C.AVG improves dramatically from 55.58\% (default) to 87.34\%, showcasing the adaptability of I\&L\&M to general-domain models. In fact, LLaVA-OV's I\&L\&M resulted in the highest RaTEScore and C.AVG, higher than CXR-LLaVA's L\&M result. These improvements can be attributed to the incorporation of clinical writing samples and grounding cues of L\&M.

For \textbf{Qwen2.5VL}, I\&L\&M yields significant gains, especially in RadGraph-XL (0.0812 vs. 0.0534) and C.AVG (74.83\% vs. 61.27\%). Similarly, \textbf{Llama3.2V} sees marked improvements in RadGraph-XL (0.1533 vs. 0.1412) and A.AVG (42.99\% vs. 41.32\%). The RadGraph-XL score, 0.1533, is actually highest among all the models and methods. These results highlight the potential of I\&L\&M to bridge the gap between general-purpose models and domain-specific tasks, making them more clinically relevant and robust.

\subsection{Is Look \& Mark really better than Look or Mark?}

\begin{figure}[thbp]
    \centering
    \includegraphics[width=1.0\linewidth]{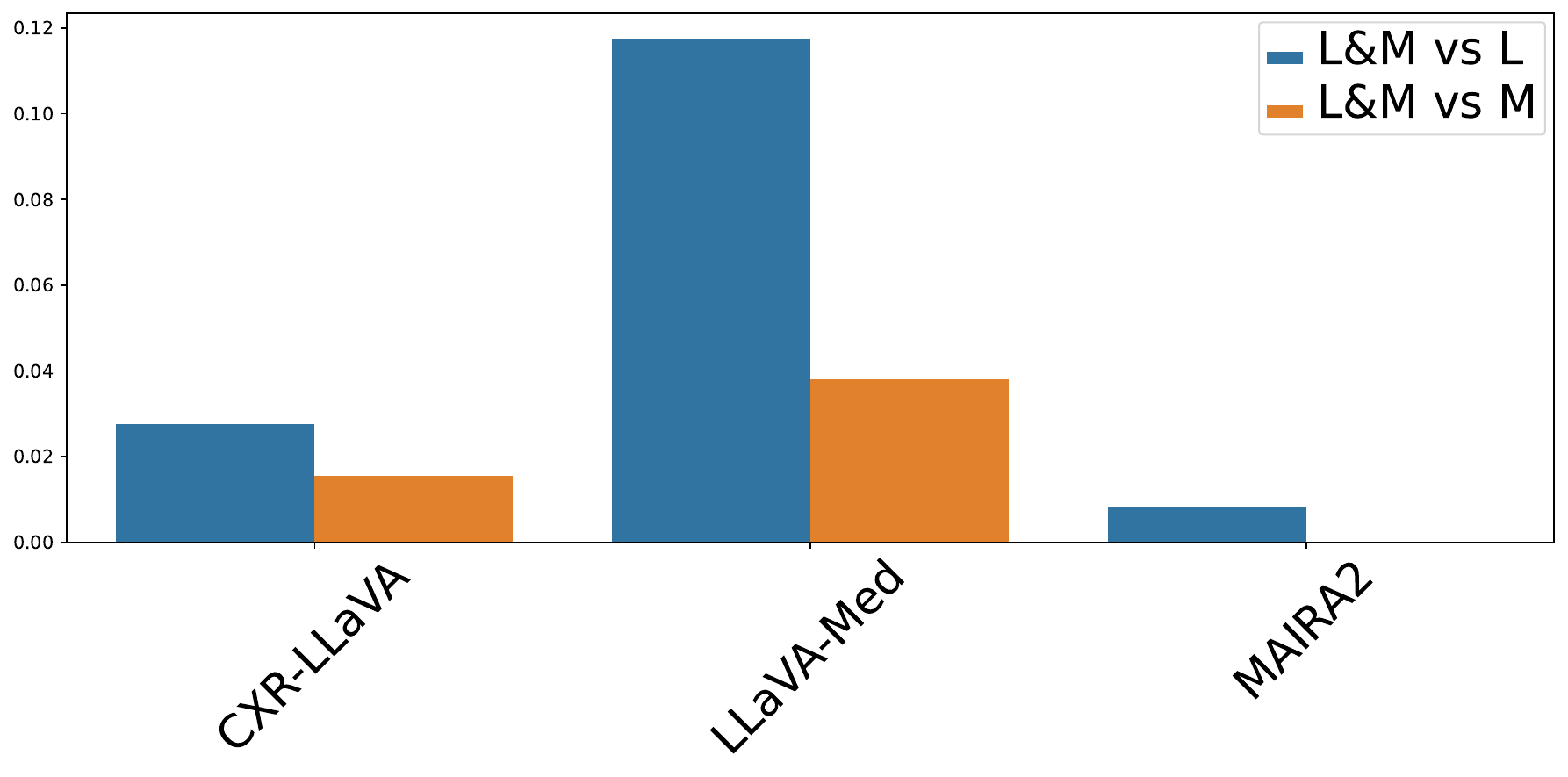}
   
    \caption{Performance increase/decrease in \textbf{A.AVG} of L\&M compared to L and M for domain-specific models.}
    \label{fig:medical_bar_plots}
\end{figure}

\begin{figure}[thbp]
    \centering
    \includegraphics[width=1.0\linewidth]{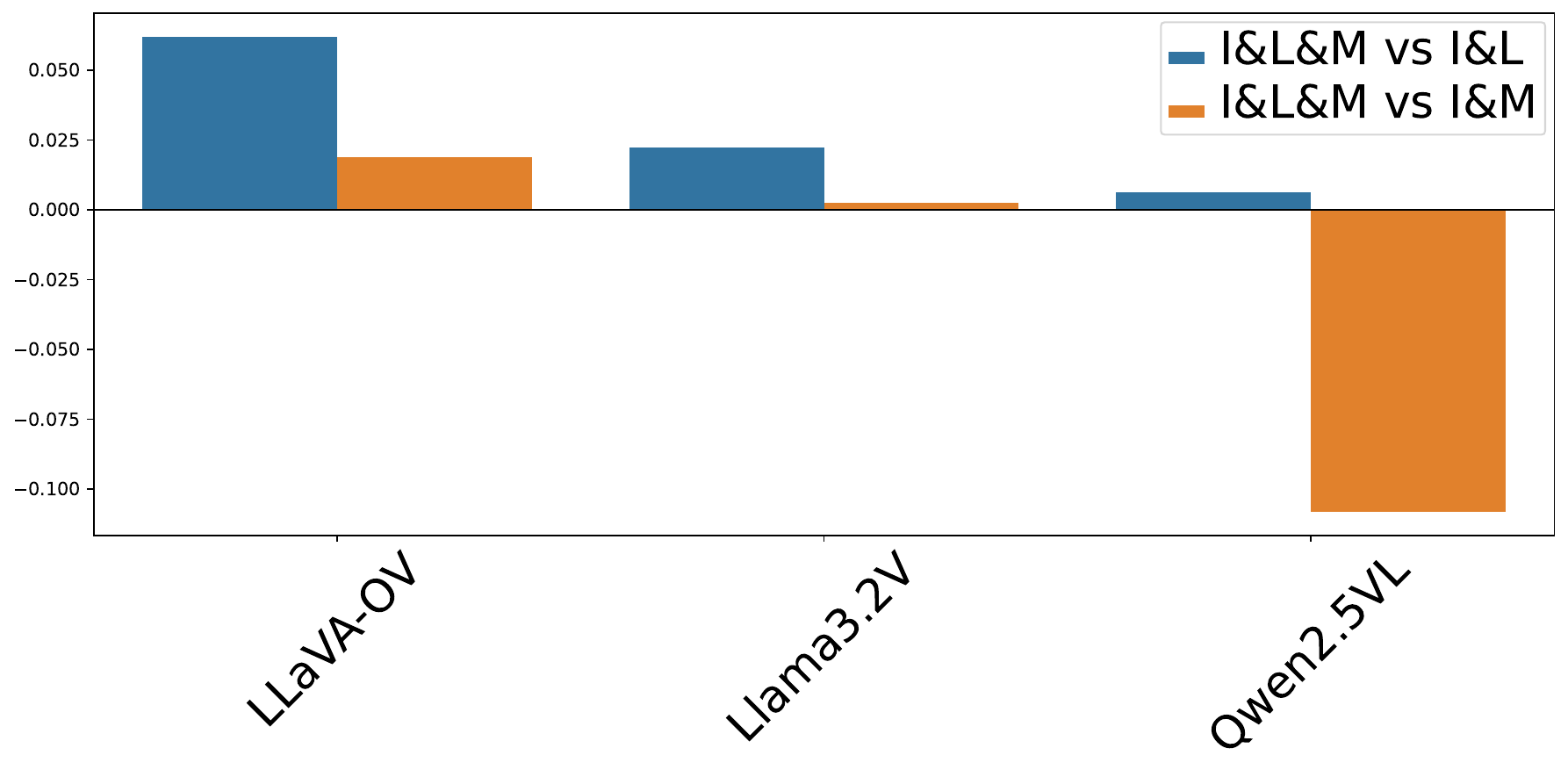}
    
    \caption{Performance increase/decrease in \textbf{A.AVG} of I\&L\&M compared to I\&L and I\&M for general-purpose models.}
    \label{fig:general_bar_plots}
\end{figure}

Figures~\ref{fig:medical_bar_plots} and~\ref{fig:general_bar_plots} analyze the relative performance of \textbf{Look \& Mark (L\&M)} compared to using only \textbf{Look (L)} or \textbf{Mark (M)}.
\subsubsection{Domain-Specific Models (Figure~\ref{fig:medical_bar_plots})}

For domain-specific models, the performance improvements achieved by combining fixation cues (L) and bounding box grounding (M) in L\&M consistently outperform using either method alone. The \textbf{CXR-LLaVA} model demonstrates significant gains with L\&M. L\&M achieves an A.AVG improvement of 1.5\% compared to M and 2.8\% compared to L. \textbf{LLaVA-Med} demonstrates the highest improvements with L\&M. L\&M achieves an A.AVG improvement of 3.8\% compared to M and 11.8\% compared to L. The \textbf{MAIRA2} model shows no performance gain with having additional fixation cues. This strengthens our finding that MAIRA2's performance depends too much on the grounded visual prompt. Still, L\&M performed better than L in all models, although very small (0.8\%) for MAIRA2. This shows that grounding can be more effective than fixation for report generation. 

\subsubsection{General-Purpose Models with In-Context Learning (Figure~\ref{fig:general_bar_plots})}

For general-purpose models, in-context learning combined with Look \& Mark (I\&L\&M) leads to significant performance gains over I\&L in all models. This also strengthens the point that grounding can be more effective than fixation for report generation. \textbf{LLaVA-OV} with I\&L\&M consistently outperforms both I\&L and I\&M, with a substantial increase of 6.2\% compared to I\&L and 1.9\% compared to I\&M. While the improvements are smaller for \textbf{Llama3.2V}, I\&L\&M still achieves a positive increase of 2.3\% compared to I\&L, highlighting incremental gains. However, the increase compared to I\&M is minimal (0.3\%), indicating that the model may struggle to fully utilize fixation cues alongside bounding boxes.
\textbf{Qwen2.5VL} with I\&L\&M outperforms I\&L with a small incarse of 0.6\% but underperforms when compared to I\&M (-10.82\%). This result also confirms that models trained for Grounding does not have capacity to effectively use fixation cues.

\subsection{Expert evaluation to confirm Look \& Mark reducing the errors or hallucinations}
To further validate the effectiveness of \textbf{Look \& Mark (L\&M)} in reducing hallucinations and clinically significant errors, we conducted an expert evaluation involving three radiologists with varied levels of experience. The goal of this evaluation was to assess whether L\&M-generated reports demonstrated fewer errors compared to reports generated by other methods. 

Three radiologists performed a blind evaluation of generated reports and annotated the number of clinically significant errors based on predefined error categories adapted from the ReXVal dataset \cite{yu2023evaluating}. The error categories were as follows:
\begin{enumerate}[noitemsep,topsep=0pt]
\item \textbf{False prediction of finding} 
\item \textbf{Omission of finding}
\item \textbf{Incorrect location/position of finding}
\item \textbf{Incorrect severity of finding}
\item \textbf{Mention of comparison that is not present}
\end{enumerate}

To assess inter-annotator reliability, we calculated Krippendorff’s Alpha for the radiologists’ annotations, which resulted in a score of 0.647 \cite{krippendorff2011content}. This indicates a moderate level of agreement, reflecting consistency in identifying clinically significant errors across the evaluated reports. Minor variability in annotations may stem from subjective differences in error interpretation.

Table~\ref{tab:expert_eval_result} summarizes the results, presenting the average number of errors per generated report across methods and models.

\begin{table}[htbp]
\centering
\begin{tabular}{lcc}
\toprule
\textbf{Models (Methods)}                        & \textbf{Errors}  \\ \midrule
CXR-LLaVA (L\&M)                       & 1.75      \\
MAIRA2 (-)               & 1.75     \\ 
LLaVA-OV (I\&L\&M)              & 1.88    \\
Qwen2.5VL (I\&L\&M)              & 2.12    \\
CXR-LLaVA (-)               & 2.18     \\ 
\bottomrule
\end{tabular}
\caption{Expert evaluation of clinically significant errors (average per generated report).}
\label{tab:expert_eval_result}
\end{table}

The results indicate that L\&M reduces clinically significant errors as \textbf{CXR-LLaVA (L\&M)} achieved the lowest average error count (1.75), while CXR-LLaVA baseline method had the highest average error count (2.18). This highlights the effectiveness of grounding fixation in reducing hallucinations and aligning reports with clinical standards. Similarly, \textbf{LLaVA-OV (I\&L\&M)} performed well, with an error count of 1.88, demonstrating the adaptability of L\&M combined with in-context learning for general-purpose models. In fact, both \textbf{LLaVA-OV (I\&L\&M)} and \textbf{Qwen2.5VL (I\&L\&M)} surpassed the baseline methods of CXR report generation model such as \textbf{CXR-LLaVA (-)}.

Figure~\ref{tab:qual_analysis} provides qualitative analysis of model outputs, comparing generated reports from different methods against ground truth reports. Three examples are shown, with clinically significant errors marked in red, as identified by radiologists. The examples include cases of pneumothorax, pleural effusion, and atelectasis. This Figure shows the effect of L\&M as CXR-LLaVA with our method significantly reduces the number of errors in the generated reports.

These findings confirm that L\&M enhances clinical accuracy while reducing hallucinations, making it a robust framework for CXR report generation across both domain-specific and general-purpose multimodal models.

\begin{figure*}[t]
\scriptsize
\centering
\begin{tabular}{p{2.3cm}|p{4cm}|p{4cm}|p{4cm}}
\toprule
 \bf Methods & \textbf{Example 1}      & \textbf{Example 2}    & \textbf{Example 3}    \\ \midrule
\multirow{7}{*}[-3pt]{\textbf{Ground Truth}}  & no pneumothorax. support apparatus. mediastinum normal. effusions, atelectasis, consolidation, or mass in the right hemithorax.     & support apparatus. no pneumothorax. mediastinum normal. bilateral pleural effusions with superimposed atelectasis versus consolidation.          & a large right pleural effusion is present. right apical pneumothorax is present. the heart is normal in size. pulmonary vessels in the left lung are normal in caliber. no acute osseous abnormality is present.    \\ \midrule

\multirow{8}{*}[-3pt]{\textbf{CXR-LLaVA (-)}}   & The chest radiograph reveals \textcolor{red}{a left-sided pneumothorax.} There is also \textcolor{red}{some improved aeration at the left base.} No other significant abnormalities are noted in the lungs, mediastinum, heart, and thorax. & The radiologic report reveals \textcolor{red}{a small left apical pneumothorax.} There is also evidence of \textcolor{red}{pneumomediastinum and subcutaneous emphysema in the left neck and chest wall.} \textcolor{red}{A small left pleural effusion} is present. \textcolor{red}{The left basal consolidation}, likely due to atelectasis, is noted.        & The radiologic report reveals a persistent moderate right-sided pleural effusion. There is \textcolor{red}{no evidence of pneumothorax.} The left lung appears clear. The cardiomediastinal silhouette shows no abnormalities. A moderate upper right-sided pneumothorax is also present.          \\ \midrule

\multirow{4}{*}[-3pt]{\textbf{CXR-LLaVA (L\&M)}}  & 
The chest radiograph reveals a right lower lobe consolidation and a right pleural effusion. The left lung appears clear. The heart size is within normal limits. There is no evidence of pneumothorax. & The chest radiograph reveals an opacity in the \textcolor{red}{left lower lobe, likely indicative of a combination of atelectasis and a small pleural effusion.} The mediastinal and cardiac contours appear normal. No pneumothorax is observed.  & The chest radiograph reveals a moderate right pleural effusion. The left lung appears clear. There is evidence of right basilar atelectasis, likely due to the pleural effusion. No focal consolidations are observed. A right apical pneumothorax is present. The cardiac silhouette is stable.             \\ \midrule

\end{tabular}
\caption{Expert analysis of model outputs. Red-colored text shows the clinically significant error marked by radiologist.}
\label{tab:qual_analysis}
\end{figure*}

\section{Conclusion}
\label{sec:conclusion}

This study introduces \textbf{Look \& Mark (L\&M)}, a novel approach to radiology report generation that integrates radiologist fixation cues (Look) with bounding box annotations (Mark) to guide multimodal Large Language Models (LLMs). By combining these complementary grounding strategies, L\&M significantly improves the clinical relevance of generated reports, reduces hallucinations, and enhances model alignment with real-world diagnostic workflows. Importantly, L\&M achieves these gains without requiring extensive fine-tuning, leveraging in-context learning to adapt both general-purpose and domain-specific models alike.

Our experiments demonstrate that L\&M significantly boosts performance across both lexical and clinical evaluation metrics, with the largest gains observed in clinical metrics such as RaTEScore and RadGraph-XL. For instance, CXR-LLaVA achieved a 1.2\% improvement in overall metrics (A.AVG) compared to baseline prompting, while LLaVA-Med demonstrated a remarkable 9.2\% boost. General-purpose models also benefited significantly, with LLaVA-OV achieving an 87.3\% clinical average (C.AVG), the highest among all tested models, even surpassing domain-specific models trained explicitly for chest X-ray report generation. Furthermore, expert radiologist evaluations confirmed the clinical reliability of L\&M, with fewer errors (by 0.43 average errors per report) in categories such as false predictions, omissions, and incorrect severity descriptions. These results highlight that grounding multimodal LLMs with both bounding boxes and fixation cues provides a synergistic effect, improving performance across diverse models and tasks.

By eliminating the need for retraining, L\&M offers a scalable and practical solution for deploying advanced AI systems in low-resource clinical environments. This makes it particularly suited for improving diagnostic workflows in settings with limited computational resources, while still achieving state-of-the-art performance.

Future work will focus on extending L\&M to other medical imaging modalities, such as CT and MRI, and exploring automated grounding for the bounding boxes of abnormalities. These advancements will further enhance L\&M's potential to become a foundational framework for reliable and scalable AI-driven diagnostics in low-resource healthcare settings.

\section*{Limitations} \label{sec:limitation}

While \textbf{Look \& Mark (L\&M)} demonstrates significant improvements in radiology report generation, several limitations remain that warrant further investigation.

First, L\&M, as implemented in this study, relies on single-view chest X-rays, whereas clinical practice often incorporates multiple views (e.g., frontal and lateral). Multi-view integration could provide a more comprehensive understanding of anatomical structures and pathologies, reducing the risk of missing findings that are evident only in specific views. Future work should extend L\&M to support multi-view training and inference to align more closely with real-world diagnostic workflows.

Second, the use of bounding box annotations and fixation data  requires expert input for dataset creation. While L\&M leverages these resources effectively, the scalability of this approach may be limited in settings where such annotations are unavailable. Exploring alternative strategies, such as weakly supervised learning or automatic fixation prediction and bounding box grounding, could reduce the reliance on expert-labeled data.

Lastly, this study focuses exclusively on chest X-rays, limiting its generalizability to other medical imaging modalities. Expanding L\&M to support other modalities, such as CT, MRI, or ultrasound, would enhance its applicability across broader radiology and clinical domains.

\section*{Broader Impacts and Ethics Statement}
\label{sec:ethics}
The development of \textbf{Look \& Mark (L\&M)} has the potential to positively transform radiology workflows by improving diagnostic accuracy and reducing errors. All data used in this research adhered to strict ethical guidelines. MIMIC-CXR and related datasets used are publicly available and contain de-identified patient information. To access MIMIC-CXR and related datasets, researchers have completed necessary training course and signed the data use agreement. 

While L\&M demonstrates significant promise, we acknowledge the potential risks associated with the deployment of AI in healthcare. These include the propagation of biases present in training datasets and the possibility of over-reliance on AI-generated reports, particularly in high-stakes clinical environments. To mitigate these risks, L\&M is explicitly designed as an assistive tool to support, rather than replace, radiologist decision-making. Additionally, future work will involve rigorous evaluation of performance across diverse populations and imaging settings to identify and mitigate biases, ensuring equity and fairness in diagnostic outcomes.

\clearpage
\bibliography{acl_latex}

\clearpage
\appendix
\section*{Appendix}
\label{sec:appendix}

The three examples shown in 
Figure~\ref{tab:qual_analysis} can be also regarded as the three exemplar reports that are used for in-context learning.

\begin{figure*}[thbp]
    \centering
    \includegraphics[width=1.0\linewidth]{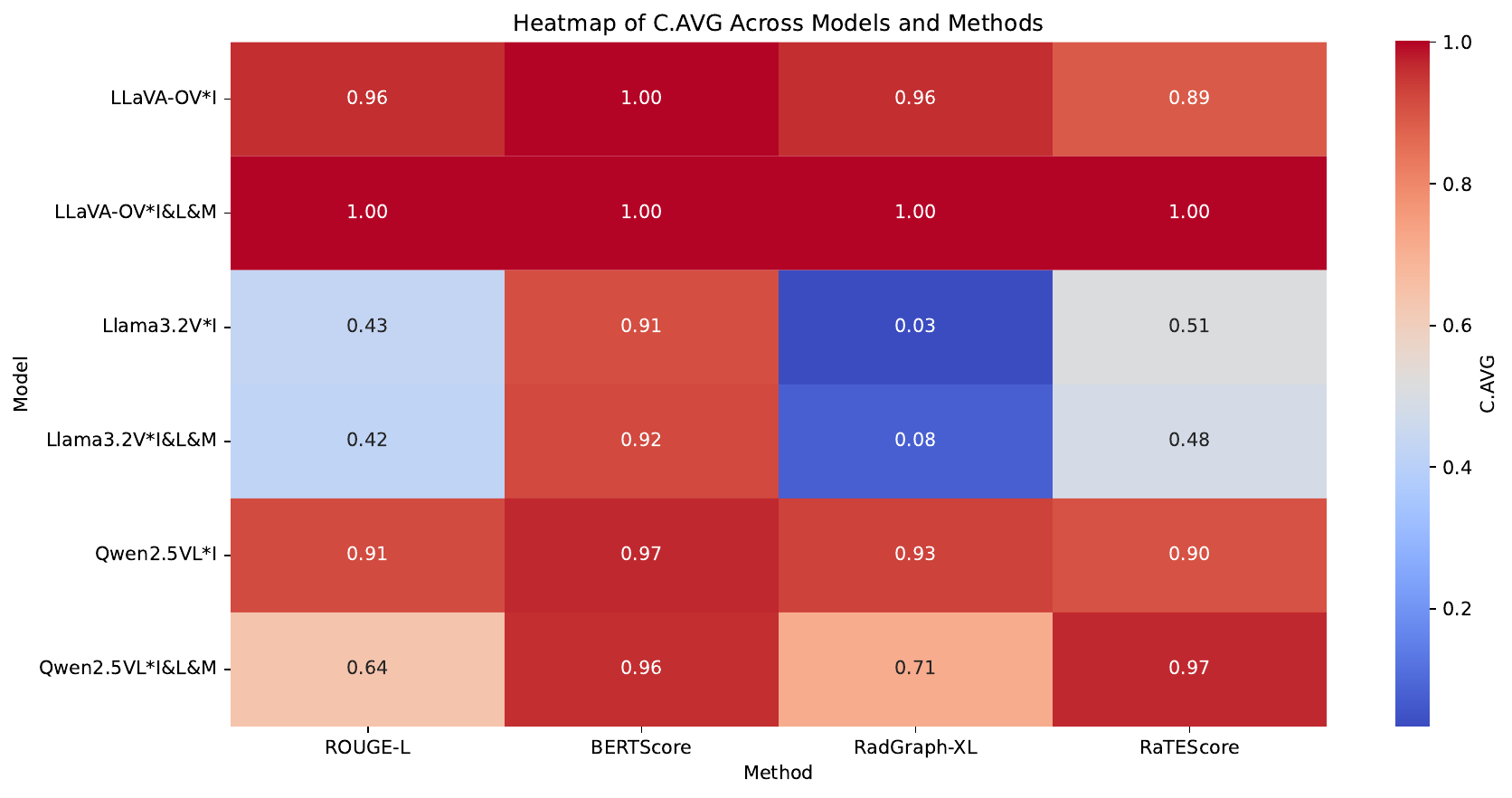}
    \caption{Heatmap of normalized scores across general-purpose models to compare in-context learning and our method.}
    \label{fig:icl_heatmap}
\end{figure*}

\begin{table*}[htbp]
\centering
\begin{tabular}{llcccccc}
\toprule
\textbf{Model} & \textbf{Method} & \textbf{RG-L} & \textbf{BERT} & \textbf{RadG} & \textbf{RaTE} & \textbf{C.AVG (\%)} & \textbf{A.AVG (\%)} \\ 
\midrule
CXR-LLaVA & - & 0.1653 & 0.8586 & 0.1107 & 0.4730 & 84.42 & 73.21 \\
CXR-LLaVA & L & 0.1652 & 0.8592 & 0.1048 & 0.4626 & 81.46 & 72.04 \\
CXR-LLaVA & M & 0.1672 & 0.8579 & 0.1093 & 0.4687 & 83.55 & 73.07 \\
\textbf{CXR-LLaVA} & \textbf{L\&M} & \textbf{0.1697} & \textbf{0.8602} & 0.1148 & 0.4752 & 86.01 & \textbf{74.40} \\
MAIRA2 & - & 0.1460 & 0.8492 & 0.0868 & 0.4507 & 74.35 & 66.70 \\
MAIRA2 & L & 0.1419 & 0.8487 & 0.0824 & 0.4460 & 72.45 & 65.44 \\
MAIRA2 & M & 0.1469 & 0.8489 & 0.0810 & 0.4574 & 73.16 & 66.31 \\
MAIRA2 & L\&M & 0.1469 & 0.8489 & 0.0810 & 0.4574 & 73.16 & 66.31 \\
LLaVA-Med & - & 0.0942 & 0.8392 & 0.0000 & 0.2445 & 24.99 & 40.62 \\
LLaVA-Med & L & 0.0818 & 0.8216 & 0.0011 & 0.2511 & 26.02 & 39.15 \\
LLaVA-Med & M & 0.0900 & 0.8281 & 0.0270 & 0.3107 & 40.56 & 46.09 \\
LLaVA-Med & L\&M & 0.0817 & 0.8253 & 0.0295 & 0.4191 & 52.46 & 49.81 \\ \midrule
Llama3.2V & - & 0.0413 & 0.7652 & 0.1412 & 0.0027 & 46.34 & 41.32 \\
Llama3.2V & L & 0.0370 & 0.7656 & 0.1155 & 0.0035 & 38.01 & 37.39 \\
Llama3.2V & M & 0.0400 & 0.7697 & 0.1528 & 0.0078 & 50.64 & 42.89 \\
Llama3.2V & L\&M & 0.0393 & 0.7694 & 0.1494 & 0.0071 & 49.44 & 42.30 \\
Llama3.2V & I & 0.0415 & 0.7652 & 0.1471 & 0.0039 & 48.38 & 42.13 \\
Llama3.2V & I\&L & 0.0374 & 0.7654 & 0.1269 & 0.0031 & 41.70 & 38.80 \\
Llama3.2V & I\&M & 0.0400 & 0.7694 & 0.1479 & 0.0074 & 49.00 & 42.23 \\
\textbf{Llama3.2V} & \textbf{I\&L\&M} & 0.0402 & 0.7696 & \textbf{0.1533} & 0.0089 & 50.91 & 42.99 \\
LLaVA-OV & - & 0.0518 & 0.8085 & 0.0471 & 0.3936 & 55.58 & 47.14 \\
LLaVA-OV & L & 0.0484 & 0.8087 & 0.0485 & 0.4029 & 57.00 & 47.31 \\
LLaVA-OV & M & 0.0565 & 0.8110 & 0.0518 & 0.4567 & 63.56 & 50.95 \\
LLaVA-OV & L\&M & 0.0527 & 0.8107 & 0.0497 & 0.4531 & 62.51 & 50.07 \\
LLaVA-OV & I & 0.0920 & 0.8384 & 0.1101 & 0.4354 & 80.41 & 62.50 \\
LLaVA-OV & I\&L & 0.0738 & 0.8268 & 0.1068 & 0.4386 & 79.67 & 59.79 \\
LLaVA-OV & I\&M & 0.0966 & 0.8350 & 0.1081 & 0.4685 & 83.13 & 64.05 \\
LLaVA-OV & I\&L\&M & 0.0959 & 0.8365 & 0.1145 & 0.4893 & 87.34 & 65.69 \\
Qwen2.5VL & - & 0.0576 & 0.8080 & 0.0534 & 0.4291 & 61.27 & 50.08 \\
Qwen2.5VL & L & 0.0530 & 0.7989 & 0.0461 & 0.3926 & 55.14 & 46.88 \\
Qwen2.5VL & M & 0.0496 & 0.7980 & 0.0588 & 0.4545 & 65.63 & 50.65 \\
Qwen2.5VL & L\&M & 0.0427 & 0.7933 & 0.0528 & 0.4488 & 63.08 & 48.71 \\
Qwen2.5VL & I & 0.0877 & 0.8104 & 0.1063 & 0.4416 & 79.80 & 61.10 \\
Qwen2.5VL & I\&L &  0.0650 & 0.8047 & 0.0812 & 0.4469 & 80.02 & 58.38 \\
\textbf{Qwen2.5VL} &\textbf{I\&M} & 0.0914 & 0.8201 & 0.1175 & \textbf{0.4914} & \textbf{88.53} & 65.26 \\
Qwen2.5VL & I\&L\&M & 0.0614 & 0.8045 & 0.0812 & 0.4730 & 74.83 & 55.88 \\
\bottomrule
\end{tabular}
\caption{Performance metrics for all models and methods. Methods include: default prompt (-), eye gaze as a heat map (L), abnormalities bounding box (M), and our method (L\&M). In-context learning, noted as I. RG-L is the acronym for ROUGE-L, BERT is the acronym for BERTScore, RadG is the acronym for RadGraph-XL, and RaTE is the acronym for RaTEScore. The best scores for each metric are bolded and the models with the scores are also bolded.}
\label{tab:all_model_metrics}
\end{table*}


Table~\ref{tab:all_model_metrics} presents the performance metrics for all models across various prompting methods, including default prompting (-), eye gaze as a heatmap (L), bounding box grounding (M), and the proposed Look \& Mark (L\&M) strategy. The evaluation also includes in-context learning (I) and its combinations with Look and Mark (e.g., I\&L, I\&M, and I\&L\&M). Key metrics include ROUGE-L (RG-L), BERTScore, RadGraph-XL (RadG), RaTEScore, Clinical Average (C.AVG), and All Metrics Average (A.AVG).

Since bounding boxes are rendered directly on top of the original chest X-ray images, one might worry that they could obscure clinically important details. Despite this potential for partial visual obstruction, our experimental results demonstrate that bounding box grounding contributes positively to both lexical and clinical performance metrics across models. This suggests that the benefit of explicit spatial localization outweighs any minor loss of visual fidelity due to overlaid markings. Notably, even general-domain models—without prior training on medical data—show improved performance when guided by these visual prompts. This underscores the effectiveness of grounding as a prompting strategy, even when visual clarity is modestly reduced.

Figure~\ref{fig:icl_heatmap} visualizes the normalized clinical average scores (C.AVG) across general-purpose models when using different prompting strategies, including in-context learning (I) and the proposed Look \& Mark method combined with in-context learning (I\&L\&M). The evaluated models include \textbf{LLaVA-OV}, \textbf{Llama3.2V}, and \textbf{Qwen2.5VL}, with metrics such as ROUGE-L, BERTScore, RadGraph-XL, and RaTEScore contributing to the C.AVG calculation. \textbf{LLaVA-OV (I\&L\&M)} achieves the highest overall C.AVG across all metrics, with normalized scores close to 1.0, showcasing the effectiveness of combining Look \& Mark with in-context learning. \textbf{Qwen2.5VL} and \textbf{Llama3.2V} show varied improvements with I\&L\&M compared to the baseline I, particularly in clinical metrics such as RaTEScore and RadGraph-XL. The variability across models indicates that the integration of Look \& Mark enhances the adaptability of general-purpose models for clinical tasks, particularly when evaluated with clinical relevance metrics.

This heatmap provides a clear comparative analysis of model performance under different prompting strategies, emphasizing the contributions of Look \& Mark in reducing hallucinations and aligning outputs with clinical standards.


\end{document}